\documentclass[letterpaper, 10 pt, conference]{ieeeconf}  

\IEEEoverridecommandlockouts                              
\let\labelindent\relax

\usepackage{enumitem}
\usepackage{amsmath} 
\usepackage{amssymb}  
\usepackage{amsfonts}

\newtheorem{remark}{Remark}[section]

\usepackage{graphicx}
\usepackage{epstopdf}

\usepackage{subfigure}
\usepackage{float} 
\usepackage[skip=0pt,font=footnotesize]{caption}

\usepackage[margin = 1in]{geometry}

\usepackage{subcaption}

\usepackage{etaremune}

\usepackage{tikz}
\usetikzlibrary{calc,patterns,decorations.pathmorphing,decorations.markings}
\usetikzlibrary{shapes,arrows}
\usepackage{verbatim}

\tikzstyle{block} = [draw, fill=blue!20, rectangle, 
    minimum height=3em, minimum width=6em]
\tikzstyle{sum} = [draw, fill=blue!20, circle, node distance=1cm]
\tikzstyle{input} = [coordinate]
\tikzstyle{output} = [coordinate]
\tikzstyle{pinstyle} = [pin edge={to-,thin,black}]

\usetikzlibrary{positioning}
\usetikzlibrary{math}

\usepackage{tikz}
\usetikzlibrary{shapes,arrows,calc,positioning}
\tikzstyle{bigblock} = [draw, fill=blue!20, rectangle, 
    minimum height=6em, minimum width=8em]
\tikzstyle{medblock} = [draw, fill=blue!20, rectangle, 
    minimum height=4em, minimum width=4em]    
\tikzstyle{mux} = [draw, fill=black!20, rectangle, 
    minimum height=5em, minimum width=0.1em]    
\tikzstyle{smallblock} = [draw, fill=blue!20, rectangle, 
    minimum height=3em, minimum width=4em]
\tikzstyle{sum} = [draw, fill=blue!20, circle, node distance=1cm]
\tikzstyle{signal} = [coordinate]
\tikzstyle{pinstyle} = [pin edge={to-,thin,black}]
\tikzstyle{block} = [draw, fill=blue!20, rectangle, 
    minimum height=3em, minimum width=6em]
\tikzstyle{blockS} = [draw, fill=blue!20, rectangle, 
    minimum height=3em, minimum width=4em]    
\tikzstyle{input} = [coordinate]
\tikzstyle{output} = [coordinate]
\usetikzlibrary{matrix}

\tikzset{add/.style n args={4}{
    minimum width=1mm,
    path picture={
        \draw[black, thick] 
            (path picture bounding box.south east) -- (path picture bounding box.north west)
            (path picture bounding box.south west) -- (path picture bounding box.north east);
        }
    }
}

\tikzset{Frame_into/.pic={
        code={\tikzset{scale=1}
        \tikzmath
            {
                \l  = 1;
                \Rc = 0.15;
            } 
        \draw [thick,->] (0, 0) -- +(\l, 0);
        \draw [thick,->] (0, 0) -- +(0, \l);
        \draw [thick, fill=white] 
			    (0,0) circle [radius=\Rc];
	    \draw [thick] ({\Rc*cos(45)}, {0\Rc*sin(45)}) 
	               -- ({\Rc*cos(225)}, {0\Rc*sin(225)});
        \draw [thick] ({\Rc*cos(135)}, {0\Rc*sin(135)}) 
	               -- ({\Rc*cos(315)}, {0\Rc*sin(315)});
  }}
}

\tikzset{Frame_outof/.pic={
        code={\tikzset{scale=1}
        \tikzmath
            {
                \l  = 1;
                \Rc = 0.15;
            } 
        \draw [thick,->] (0, 0) -- +(\l, 0);
        \draw [thick,->] (0, 0) -- +(0, \l);
        \draw [thick, fill=white] 
			    (0,0) circle [radius=\Rc];
	    \draw [thick, fill=black] 
			    (0,0) circle [radius={0.2*\Rc}];
  }}
}

\usepackage{hyperref}
\usepackage{xcolor}
\hypersetup{
    colorlinks,
    linkcolor={blue!100!black},
    citecolor={blue!50!black},
    urlcolor={blue!80!black}
}

\newcommand{\bc}{\begin{center}}
\newcommand{\ec}{\end{center}}
\newcommand{\benum}{\begin{enumerate}}
\newcommand{\eenum}{\end{enumerate}}
\newcommand{\nn}{\nonumber}
\newcommand{\matl}{\left[ \begin{array}}
\newcommand{\matr}{\end{array} \right]}
\newcommand{\matls}{\left[ \begin{smallmatrix}}
\newcommand{\matrs}{\end{smallmatrix} \right]}
\newcommand{\isdef}{\stackrel{\triangle}{=}}

\newcommand{\vect}[1]{\overset{\rightharpoonup}{#1}}

\newcommand{\tr}{{\rm tr}\,}

\newcommand{\rmT}{{\rm T}}

\newcommand{\rmd}{{\rm d}}

\newcommand{\rms}{{\rm s}}

\newcommand{\BBE}{{\mathbb E}}

\newcommand{\BBR}{{\mathbb R}}

\newcommand{\SD}{{\mathcal D}}

\newcommand{\SF}{{\mathcal F}}

\newcommand{\SL}{{\mathcal L}}
\newcommand{\SM}{{\mathcal M}}
\newcommand{\SN}{{\mathcal N}}

\newcommand{\SU}{{\mathcal U}}

\renewcommand{\matl}{\begin{bmatrix}}
\renewcommand{\matr}{\end{bmatrix} }

\usepackage{multicol,lipsum}
\usepackage[export]{adjustbox}

\tikzstyle{rectRound} = [rectangle, rounded corners, text centered, draw=black, minimum width=2em]
\tikzstyle{rect} = [rectangle, text centered, draw=black, minimum width=2em]
\tikzstyle{clear} = [draw=none, text centered, minimum width=2em]
\tikzstyle{circ} = [circle, text centered, draw=black, minimum width = 2em]
\tikzstyle{circClear} = [circle, text centered, minimum width = 2em]
\tikzstyle{arrow} = [thick, -latex]
\makeatletter
\DeclareRobustCommand{\rvdots}{%
  \vbox{
    \baselineskip4\p@\lineskiplimit\z@
    \kern-\p@
    \hbox{.}\hbox{.}\hbox{.}
  }}
\makeatother

\usepackage{comment}

\usepackage{tikz}
\usetikzlibrary{calc,patterns,decorations.pathmorphing,decorations.markings}
\usetikzlibrary{shapes,arrows}
\usetikzlibrary{positioning}
\usetikzlibrary{math}
\usepackage{verbatim}

\tikzstyle{block} = [draw, fill=blue!20, rectangle, 
    minimum height=3em, minimum width=6em]
\tikzstyle{sum} = [draw, fill=blue!20, circle, node distance=1cm]
\tikzstyle{input} = [coordinate]
\tikzstyle{output} = [coordinate]
\tikzstyle{pinstyle} = [pin edge={to-,thin,black}]
\tikzset{add/.style n args={4}{
    minimum width=2mm,
    path picture={
        \draw[black, thick] 
            (path picture bounding box.south east) -- (path picture bounding box.north west)
            (path picture bounding box.south west) -- (path picture bounding box.north east);
        }
    }
}

\tikzset{
        partial ellipse/.style args={#1:#2:#3}{
            insert path={+ (#1:#3) arc (#1:#2:#3)}
        }
    }
\tikzset{Frame_into/.pic={
        code={\tikzset{scale=1}
        \tikzmath
            {
                \l  = 1;
                \Rc = 0.15;
            } 
        \draw [thick,->] (0, 0) -- +(\l, 0);
        \draw [thick,->] (0, 0) -- +(0, \l);
        \draw [thick, fill=white] 
			    (0,0) circle [radius=\Rc];
	    \draw [thick] ({\Rc*cos(45)}, {0\Rc*sin(45)}) 
	               -- ({\Rc*cos(225)}, {0\Rc*sin(225)});
        \draw [thick] ({\Rc*cos(135)}, {0\Rc*sin(135)}) 
	               -- ({\Rc*cos(315)}, {0\Rc*sin(315)});
  }}
}

\tikzset{Frame_outof/.pic={
        code={\tikzset{scale=1}
        \tikzmath
            {
                \l  = 1;
                \Rc = 0.15;
            } 
        \draw [thick,->] (0, 0) -- +(\l, 0);
        \draw [thick,->] (0, 0) -- +(0, \l);
        \draw [thick, fill=white] 
			    (0,0) circle [radius=\Rc];
	    \draw [thick, fill=black] 
			    (0,0) circle [radius={0.2*\Rc}];
  }}
}

\newcommand\spiral{}
\def\spiral[#1](#2)(#3:#4:#5){
\pgfmathsetmacro{\domain}{pi*#3/180+#4*2*pi}
\draw [#1,
       shift={(#2)},
       domain=0:\domain,
       variable=\t,
       smooth,
       samples=int(\domain/0.08)] plot ({\t r}: {#5*\t/\domain})
}

\tikzstyle{spring}=[thick,decorate,decoration={zigzag,pre length=1cm, post length=1cm,segment length=6}]

\usepackage[style=ieee ,sorting=none,giveninits]{biblatex}
\usepackage{graphics} 
\usepackage{amsmath} 
\usepackage{amssymb}  

\title{\LARGE \bf
Longitudinal Flight Dynamics Control Based on Feedback Linearization and Normal Canonical Form
}

\title{\LARGE \bf
Data-driven, Model-free, Coordinate-free,  Adaptive Attitude Estimation
}

\title{Multiplicative, Retrospective Cost Adaptive Attitude Estimation}
\title{Multiplicative Adaptive Attitude Estimation with Unknown Gyro Bias}
\title{Data-driven Attitude Filtering with Unknown Gyro Bias}
\title{Retrospective Cost Attitude Filtering \\ with Noisy Measurements and Unknown Gyro Bias}

\title{Improving Long term Accuracy of Neural Network Predictions}
\title{Neural Network-Based Dynamic System Modeling with Extended Kalman Filter for Improved Long-Term Predictions}

\title{Neural filtering for Long-term State Prediction}

\title{Neural filtering for Neural Network-based Models of Dynamic Systems}

\title{A Novel Neural Filter to Improve Accuracy of \\ Neural Network Models of Dynamic Systems}

\author{Parham Oveissi, Turibius Rozario, Ankit Goel%
\thanks{Parham Oveissi is a graduate student in the Department of Mechanical Engineering, University of Maryland, Baltimore County, 1000 Hilltop Circle, Baltimore, MD 21250. {\tt\small parhamo1@umbc.edu}}%
\thanks{Turibius Rozario is an undergraduate student and a Meyerhoff Scholar in the Department of Mechanical Engineering, University of Maryland, Baltimore County, 1000 Hilltop Circle, Baltimore, MD 21250. {\tt \small s175@umbc.edu}}%
\thanks{Ankit Goel is an Assistant Professor in the Department of Mechanical Engineering, University of Maryland, Baltimore County,1000 Hilltop Circle, Baltimore, MD 21250. {\tt\small ankgoel@umbc.edu }}%
}
\usepackage{graphicx}      

\usepackage{textcomp}
\usepackage{calc}
\usepackage{mathtools}
\usepackage{xparse}
\usepackage{amssymb}
\usepackage{amsmath}
\usepackage{cases}
\usepackage{mathtools}
\usepackage{cuted}

\begin{document}

\maketitle

\begin{abstract}
The application of neural networks in modeling dynamic systems has become prominent due to their ability to estimate complex nonlinear functions.
Despite their effectiveness, neural networks face challenges in long-term predictions, where the prediction error diverges over time, thus degrading their accuracy.
This paper presents a \textit{neural filter}
to enhance the accuracy of long-term state predictions of neural network-based models of dynamic systems.
Motivated by the extended Kalman filter, the neural filter combines the neural network state predictions with the measurements from the physical system to improve the estimated state's accuracy.  
%
%
The neural filter's improvements in prediction accuracy are demonstrated through applications to four nonlinear dynamical systems. 
Numerical experiments show that the neural filter significantly improves prediction accuracy and bounds the state estimate covariance, outperforming the neural network predictions.
Furthermore, it is also shown that the accuracy of a poorly trained neural network model can be improved to the same level as that of an adequately  trained neural network model, potentially decreasing the training cost and required data to train a neural network.

\end{abstract}






\textit{\bf keywords:} neural networks, dynamical systems, 
neural network modeling, state estimation.



\section{Introduction}



The use of neural networks has surged due to advances in computational power and extensive research within the machine learning community.
Neural networks, with their inherently nonlinear activation functions, are adept at estimating complex nonlinear functions \cite{song2023approximation, higgins2021generalizing}, making them highly suitable for modeling dynamic systems and predicting future states. 
These capabilities have broad applications, such as modeling nonlinear oscillators \cite{masri1993identification}, controlling nonlinear dynamical systems \cite{narendra1992neural, chee2024performance}, predicting weather phenomena \cite{lin2023spherical}, and enhancing medical diagnostics \cite{yasnitsky2022capabilities}. 
Modeling dynamic systems is crucial across various engineering and scientific disciplines. 
Neural networks offer a powerful alternative to traditional parametric methods by learning complex system behaviors directly from data.
For instance, a procedure developed for identifying nonlinear dynamic systems using artificial neural networks demonstrated the capability to effectively predict the response of a damped Duffing oscillator under varying excitations, highlighting the neural network's high-fidelity modeling potential \cite{masri1993identification}. 
\cite{narendra1992neural} introduced models for nonlinear dynamical systems identification and control, emphasizing the feasibility of neural networks in handling more complex systems and introducing a combined linear-neural network controller structure. 
More recently, universal approximation property of neural networks has been leveraged to learn functions from small datasets.
A significant reduction in generalization error and high-order error convergence in dynamic systems and PDEs was reported in \cite{lu2019deeponet}. 
Physics-informed neural networks (PINNs) that integrate physical laws into neural network training to ensure that the models adhere to underlying physical principles have also been explored \cite{raissi2019physics}.


Despite neural networks' impressive capabilities, a significant challenge arises in the context of long-term predictions, where the error between the true system behavior and the neural network approximation tends to accumulate over time.
To address the issue of long-term prediction errors, various approaches have been proposed.
A technique based on regularization to improve robustness and reduce error accumulation was investigated in \cite{pan2018long}.
Recurrent neural networks have been investigated to capture temporal dependencies to reduce error accumulation for both interpolation and extrapolation tasks \cite{michalowska2023neural,mohajerin2019multistep}. 



The key reason for the prediction error's divergence is that the design and training of the neural network-based model do not stabilize the state error dynamics. 
Motivated by the Kalman filter \cite{chui2017kalman}, where a feedback signal from the physical system stabilizes the error dynamics, this paper presents a \textit{neural filter} to arrest the divergence of the error in the neural network state predictions. 
The main contribution of this work is thus the extension of the extended Kalman filter \cite{ribeiro2004kalman, mirtaba2023design} to the neural filter and the demonstration of the improvement in the long-term accuracy of the neural filter state predictions.  


The paper is organized as follows. 
Section \ref{sec: Neural Filter} formulates the state-estimation problem and presents the neural filter.
Section \ref{sec: Case Studies} applies the neural filter to estimate states in four typical dynamical systems. 
Finally, Section \ref{sec:Conclusions} summarizes the paper.

\section{Problem Formulation}
\label{sec: Neural Filter}
This section formulates notation and terminology associated with the state-estimation problem and presents the \textit{neural filter}. 
Consider the nonlinear system
\begin{align}
    x_{k+1} &= f \big(x_k , u_k \big) + w_k, \label{eq:linear_system_state_eqution}\\ 
    y_k &= g \big(x_k \big) + v_k, \label{eq:linear_system_measurement_eqution}
\end{align}
where, for all $k \ge 0$,
$x_k$ is the state, 
$u_k$ is the input to the system, 
$y_k$ is the measured output of the system, 
$f, g,$ are real-valued vector
functions, $w_k \sim \SN(0, Q_k)$ is the process noise, $v_k \sim \SN(0, R_k)$ is the measurement noise, and $Q_k$ and $R_k$ are process and measurement covariance matrices, respectively. 
The objective is to propagate the system states $x_k$ by approximating the function $f$ with a neural network, and then correct these predictions using the available measurements $y_k$ from the system.
In this work, we assume that the function $g$ is known.

\subsection{Neural Network Model}
Consider a dynamic system
\begin{align}
    \lambda x(t) = \SF(x(t),u(t)),
    \label{eq:ss_form}
\end{align}
where $x $ and $ u$ are the state of the system and the input to the system, respectively, $\SF$ is the dynamics of the system, and the operator $\lambda$ is either a derivative or a forward-shift operator, that is, in continuous-time systems, $\lambda x= \dot x $ and, in discrete-time systems, $\lambda x(t)= x(t+1). $ 
The objective is approximate the state of \eqref{eq:ss_form} with a discrete dynamic system
\begin{align}
    \hat x_{k+1} = NN(\hat x_k, u_k),
    \label{eq:NN_approx}
\end{align}
where $\hat x_k$ is the estimated state of the system and $NN$ denotes a neural-network approximation of the system, which is constructed as shown below. 
Note that $k$ is the iteration number and is related to continuous time $t$ as $t = k T_\rms,$ where $T_\rms$ is the discretization timestep.

To generate the training data to construct the neural-network model, we solve \eqref{eq:ss_form} from $t=0$ to $t=T_\rms.$
%
For example, in the case of continuous time system, note that 
\begin{align}
    x(T_\rms) 
        =
            x(0)
            +
            \int_0^{T_\rms} \SF(x(\tau), u(\tau)) \rmd \tau.
    \label{eq:x_t_SP1}
\end{align}
The training data consists of randomly generated samples of $x(0)$ and $u(0)$ and the corresponding $x(T_\rms),$ computed using \eqref{eq:x_t_SP1}.
In this work, we use MATLAB's \href{https://www.mathworks.com/help/matlab/ref/ode45.html}{ode45} routine to compute $x(T_\rms).$
The trained neural network, denoted by $NN(x),$ can thus be used to propagate the state at a time instant $t$ to $t+T_\rms$ using \eqref{eq:NN_approx}.

In this work, we use the MATLAB routine \href{https://www.mathworks.com/help/deeplearning/ref/dlnetwork.html#d126e70205}{dlnetwork} to define the neural network architecture.
To train the neural-network model $NN$ with the training data, which includes the input data $\{ x(0), u(0) \}$ and the output data $\{ x(T) \},$ we employ the Adam optimizer using the MATLAB routine \href{https://www.mathworks.com/help/deeplearning/ref/adamupdate.html}{adamupdate} in a mini-batch training setup. A small portion of the training data is reserved as a validation dataset throughout the training process.

\subsection{Neural Filter}
Let $NN(x, u)$ be a neural network approximation of the function $f.$
Then, the \textit{neural filter} is 
\begin{align}
    \hat x_{k+1|k} 
        &=
            NN \big(\hat x_{k|k} , u_k \big), \label{eq: prior_estimate_non-lin_sys}\\
    \hat x_{k+1|k+1} 
        &=
            \hat x_{k+1|k} + K_{k+1}\big (y_{k+1} -g (\hat x_{k+1|k})\big ), \label{eq: posterior_estimate_linsys}
\end{align}
where $\hat x_{k+1|k}$ is the \textit{prior estimate} at step $k+1$, $\hat x_{k+1|k+1}$ is the \textit{posterior estimate} at step $k+1$, and the \textit{neural filtering} correction gain $K_{k+1}$ is given by
\begin{align}
    K_{k+1} 
        =
            P_{k+1|k} C_{k+1}^{\rmT} (C_{k+1}P_{k+1|k} C_{k+1}^{\rmT} + R_{k+1})^{-1},
    \label{eq: kalman_gain_linsys}
\end{align}
where the \textit{prior covariance} matrix $P_{k+1|k}$ is given by
\begin{align}
    P_{k+1|k} &= A_k P_{k|k} A_k^\rmT + Q_k, \label{eq: prior_cov_linsys}
\end{align}
and the state transition matrix $A_k$ and the measurement matrix $C_k$ are given by 
\begin{align}
    A_k 
        &=
            \frac{\partial NN}{\partial  x} 
            \Bigg \vert _{\hat x_{k|k}, u_k}, \quad
    C_k 
        =
            \frac{\partial g}{\partial  x} 
            \Bigg \vert _{\hat x_{k|k}, u_k}.
\end{align}
The computation of the Jacobian of a neural network is described in Appendix \ref{appdx:jacobian_NN}.

%
%
Finally, the \textit{posterior covariance} matrix $P_{k+1|k+1}$ is given by
\begin{align}
    P_{k+1|k+1} 
        &=
            (I-K_{k+1}C_{k+1}) P_{k+1|k} 
            \label{eq: posterior_cov_linsys}
\end{align}

\begin{remark}
    Note that, since the \textit{neural filter}  is motivated by the extended Kalman filter, the matrices $P_{k+1|k}$ and $P_{k+1|k+1}$ are similar to the \textit{prior} and \textit{posterior covariance} matrices in EKF. 
    Therefore, 
    \begin{align}
        P_{k+1|k} &\approx \mathbb{E}[e_{k+1|k} e_{k+1|k}^\rmT],\\
        P_{k+1|k+1} &\approx \mathbb{E}[e_{k+1|k+1} e_{k+1|k+1}^\rmT],
    \end{align}
    where the \textit{prior error} $e_{k+1|k}$ and the  \textit{posterior error} $e_{k+1|k+1}$ are defined as
    \begin{align}
        e_{k+1|k} &\isdef x_{k+1} - \hat x_{k+1|k}, \\
        e_{k+1|k+1} &\isdef x_{k+1} - \hat x_{k+1|k+1}.
    \end{align}
\end{remark}


%
\begin{remark}
    Since $P_{k|k} $ is the covariance of the posterior state-estimation error, that is, $\BBE [e_{k|k} e_{k|k}^\rmT,] $ the trace of $P_{k|k}$ is approximately equal to the square of the 2-norm of the posterior state-estimation error, that is, $\BBE [e_{k|k}^\rmT e_{k|k}.] $
\end{remark}

\begin{figure}[ht]
	\centering
 		\begin{tikzpicture}[auto, node distance=2cm,>=latex']
            \scriptsize
            \small
            \draw[draw=black, fill=yellow!10] (-2.5,-4.55) 
             node [xshift=80, yshift=5pt] {\textbf{Neural Filter}} 
             rectangle ++(6.25,3.8);

             \node at (-3.5,0) (input) {$u_k$};
             
			\node at (-1,0) [ultra thick, block, align=center, node distance=0 cm, minimum width=2em] (MainSystem) {Physical\\System\\ \eqref{eq:linear_system_measurement_eqution}};
			
			\node at (-1,-1.5) [ultra thick, block, align=center, fill=green!20, minimum width=2em](NN_model) {Neural network\\Model \eqref{eq: prior_estimate_non-lin_sys}}; 

            \node at (2,-1.5) [ultra thick, block, align=center, fill=green!20, minimum width=2em](NN_Output) {Output\\ Model \eqref{eq:linear_system_measurement_eqution}}; 
			
			\node at (-1,-3.5) [ultra thick, block, align=center, node distance = 2 cm,fill=red!20, minimum width=2em](correctorSystem) {Neural Filter\\Correction \eqref{eq: posterior_estimate_linsys}};
						
			\node [ultra thick, sum, align=center, right of = NN_Output, node distance = 1.5 cm] (sum) {}; 
   
            \node at (0.75,-2.5) [ultra thick, sum, align=center, node distance = 1.5 cm](sum2) {};

			

            \draw [ultra thick,->] (input.east) |-  (MainSystem.west) ;

            \draw [ultra thick,->] ([xshift=.5cm, yshift=0cm]input.east) |-  (NN_model.180) ;
			
			\draw [ultra thick,->] (MainSystem.east) node [xshift=1em, yshift=8pt] {$y_k$} 
			        -|  (sum.north) ;
			\draw [ultra thick,->] (NN_model.east) node [xshift=0.5cm, yshift=8pt] {$\hat x_{k+1|k}$} 
			        --  (NN_Output.west) ;

            \draw [ultra thick,->] (NN_model.east) 
			        -|  (sum2.north) ;

            \draw [ultra thick, color=blue, ->] (sum2) -- +(1,0) node [xshift=2.5em, yshift=0pt] {$\hat x_{k+1|k+1}$};
           
            \draw [ultra thick,->] (NN_Output.east) node [xshift=1em, yshift=8pt] {$\hat{y}_k$} 
			        --  (sum.west) ;
           
			\draw [ultra thick,->] (sum.south) |-  (correctorSystem.east) node [xshift=.5cm, yshift=8pt] {} ;

            \draw [ultra thick,->] (correctorSystem) |-  (sum2.180);

            \draw [ultra thick,->] (NN_model) -- +(1.25,0) |- +(-0.4, -0.75) -| (correctorSystem.140);




		\end{tikzpicture}
    \caption{{Neural filter} architecture.
    }
	\label{Fig.correctorSystem_formulation}
\end{figure}
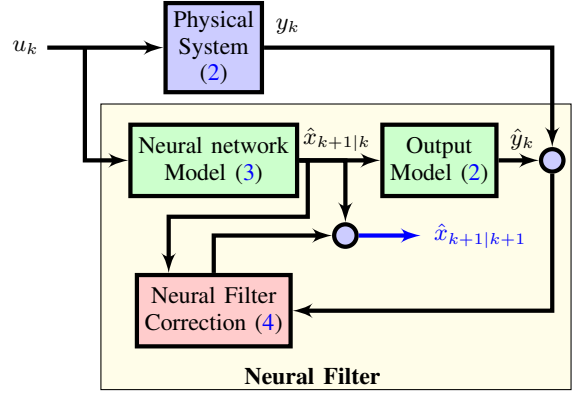

Figure \ref{Fig.correctorSystem_formulation} shows the {neural filter} architecture, which incorporates the available measurements from the physical system to improve the prediction accuracy of the neural network-based model of the physical system.

\section{Numerical Experiments}
\label{sec: Case Studies}

This section presents several case studies to demonstrate the performance of the neural filter. 
In particular, we consider the simple pendulum, the Van der Pol oscillator, the Lorenz system, and the double pendulum system to demonstrate that the neural filter maintains prediction accuracy over a long horizon. 
Note that the Lorenz system and the double pendulum system are chaotic, making long-term predictions especially challenging due to their sensitivity to initial conditions, where small changes can lead to vastly different behaviors \cite{ditto1995principles}.


\subsection{Simple Pendulum}

Consider the simple pendulum
\begin{align}
    \ell \ddot \theta + g \sin \theta =
    0.
    \label{eq:simple_pendulum}
\end{align}
Defining
$
    x
        \isdef
            \matl
                \theta & 
                \dot \theta
            \matr^\rmT, 
$
the simple pendulum \eqref{eq:simple_pendulum} can be written in the state-space form \eqref{eq:ss_form}, where 
\begin{align}
    f(x)
        \isdef
            \matl 
                x_2 \\
                -\frac{g}{\ell} \sin(x_1) 
            \matr.
\end{align}
The output is assumed to be a noisy measurement of the angle $\theta$, that is, 
\begin{align}
    y_k = \begin{bmatrix}
       1  & 0
    \end{bmatrix}x_k + v_k, 
\end{align}
where $v_k \sim \SN(0, \sigma_v^2 I_2)$ represents zero-mean Gaussian noise. 
In this example, we set $\sigma_v = 0.1$.

The training data consists of 15,000 samples of $x(0) \in \matl \SU[-\pi/2, \pi/2] \\ \SU[-5, 5] \matr.$
Using MATLAB's \href{https://www.mathworks.com/help/matlab/ref/ode45.html}{ode45} routine, $x(0.1)$ is computed according to \eqref{eq:x_t_SP1}.
In this work, we use 80 $\%$ of the data to train the model, and the remaining 20 $\%$ are used for validation during the training process.

\begin{figure}[ht]
	\centering
        \begin{tikzpicture}[scale=1.2, transform shape]

            \node[circle, draw, minimum size=0.3cm] (I-1) at (0,2) {};
            \node[anchor=east] at (I-1.west) {$x_1(0)$};
            
            \node[circle, draw, minimum size=0.3cm] (I-2) at (0,0) {};
            \node[anchor=east] at (I-2.west) {$x_2(0)$};
            
            \foreach \i in {1,...,10} {
                \node[circle, draw, minimum size=0.3cm] (H-\i) at (2,3.75-\i*0.5) {};
            }
            
            \node[circle, draw, minimum size=0.3cm] (O-1) at (4,2) {};
            \node[anchor=west] at (O-1.east) {$x_1(0.1)$};
            
            \node[circle, draw, minimum size=0.3cm] (O-2) at (4,0) {};
            \node[anchor=west] at (O-2.east) {$x_2(0.1)$};
            
            \foreach \i in {1,2} {
                \foreach \j in {1,...,10} {
                    \draw[->] (I-\i) -- (H-\j);
                }
            }
            
            \foreach \i in {1,...,10} {
                \draw[->] (H-\i) -- (O-1);
                \draw[->] (H-\i) -- (O-2);
            }

\end{tikzpicture}
 		
    \caption{Neural network architecture used to approximate the simple pendulum system.}
    \label{fig:simple pendulum NN structure}
\end{figure}
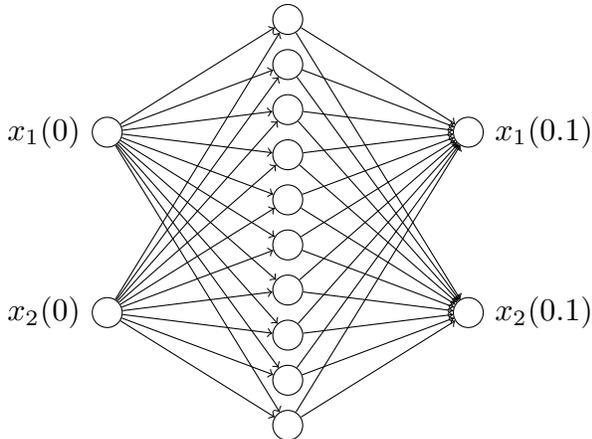

\begin{figure}[h]
    \centering
    \includegraphics[width=\columnwidth]{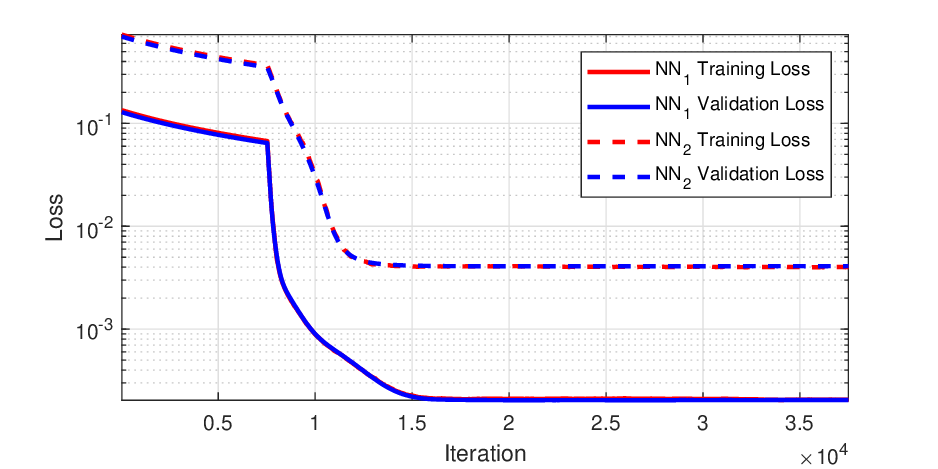}
    \caption{\textbf{Simple pendulum}. Training and validation loss on a logarithmic scale to approximate the simple pendulum system.}
    \label{fig:Pendulum_training_process_NN1_NN2}
\end{figure}

The neural network architecture used to approximate the simple pendulum is shown in Figure \ref{fig:simple pendulum NN structure}.
In particular, the neural network $\rm{NN}_1$ consists of an input layer with a dimension of 2, a single hidden layer containing 10 neurons, and an output layer with a dimension of 2. 
The hidden layer uses the rectified linear unit (ReLU) activation function, while the output layer uses a linear activation function. 
The Adam optimizer is used for training the neural network.
Training is performed with a batch size of 32, and validation is conducted every 30 iterations.
To assess the robustness of the neural filter against variations in neural network complexity, we also consider a significantly simpler architecture $\rm{NN}_2$. This network consists of an input layer with a dimension of 2, a single hidden layer with 2 neurons, and an output layer with a dimension of 2. The hidden layer employs the ReLU activation function, while the output layer utilizes a linear activation function.
Figure \ref{fig:Pendulum_training_process_NN1_NN2} shows the smoothed training and validation loss on a logarithmic scale during the training process for both trained networks $\rm{NN}_1$ and $\rm{NN}_2$.

\begin{figure}[h]
    \centering
    \includegraphics[width=\columnwidth, trim={0.5cm 0 0.8cm 0},clip]{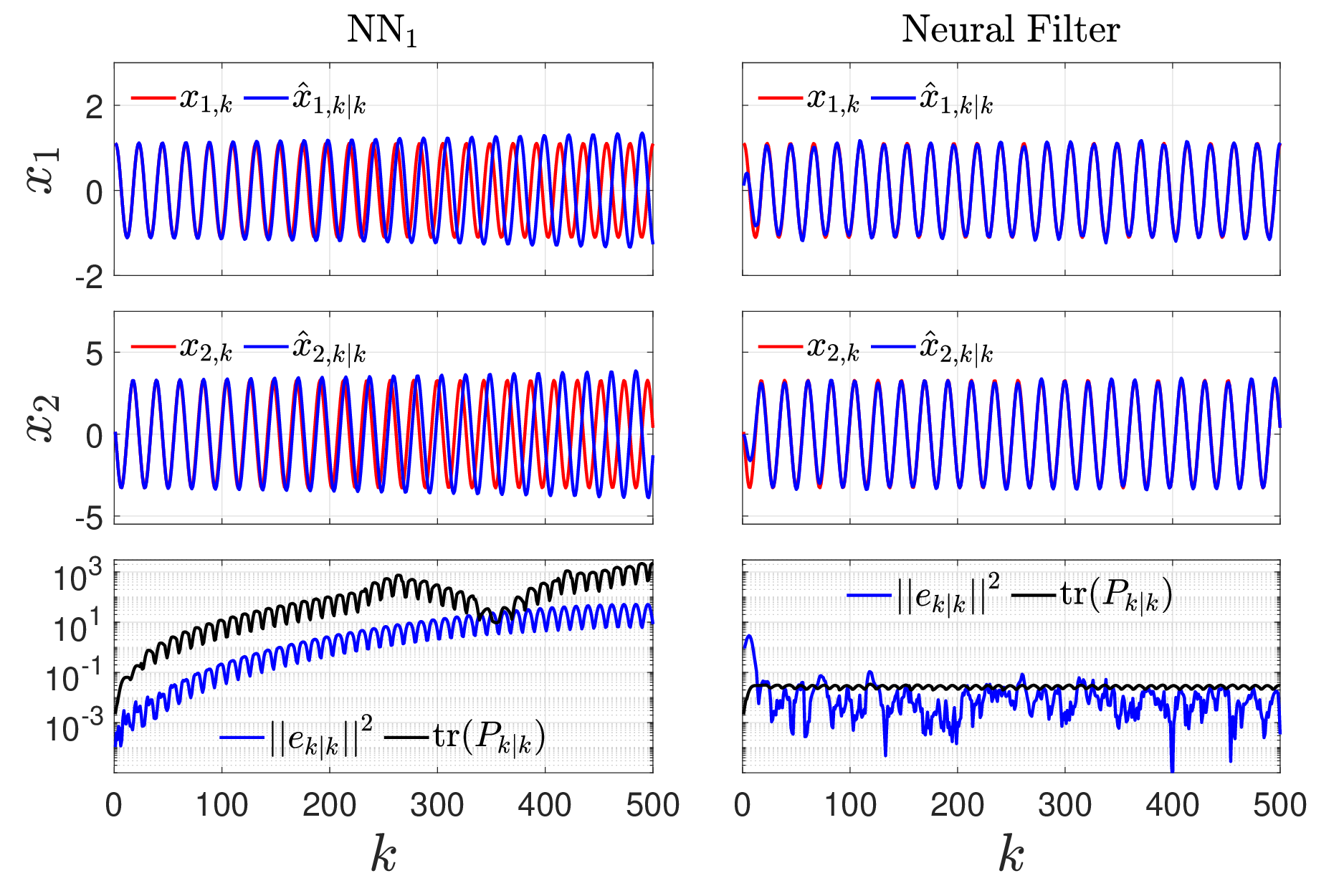}
    \caption{\textbf{Simple pendulum.} State predictions using the neural network and the neural filter using $\rm{NN}_1$. 
    }
    \label{fig:Pendulum_NN_NNEKF_fEKF_v2}
\end{figure}

\begin{figure}[h]
    \centering
    \includegraphics[width=\columnwidth, trim={0.5cm 0 0.8cm 0},clip]{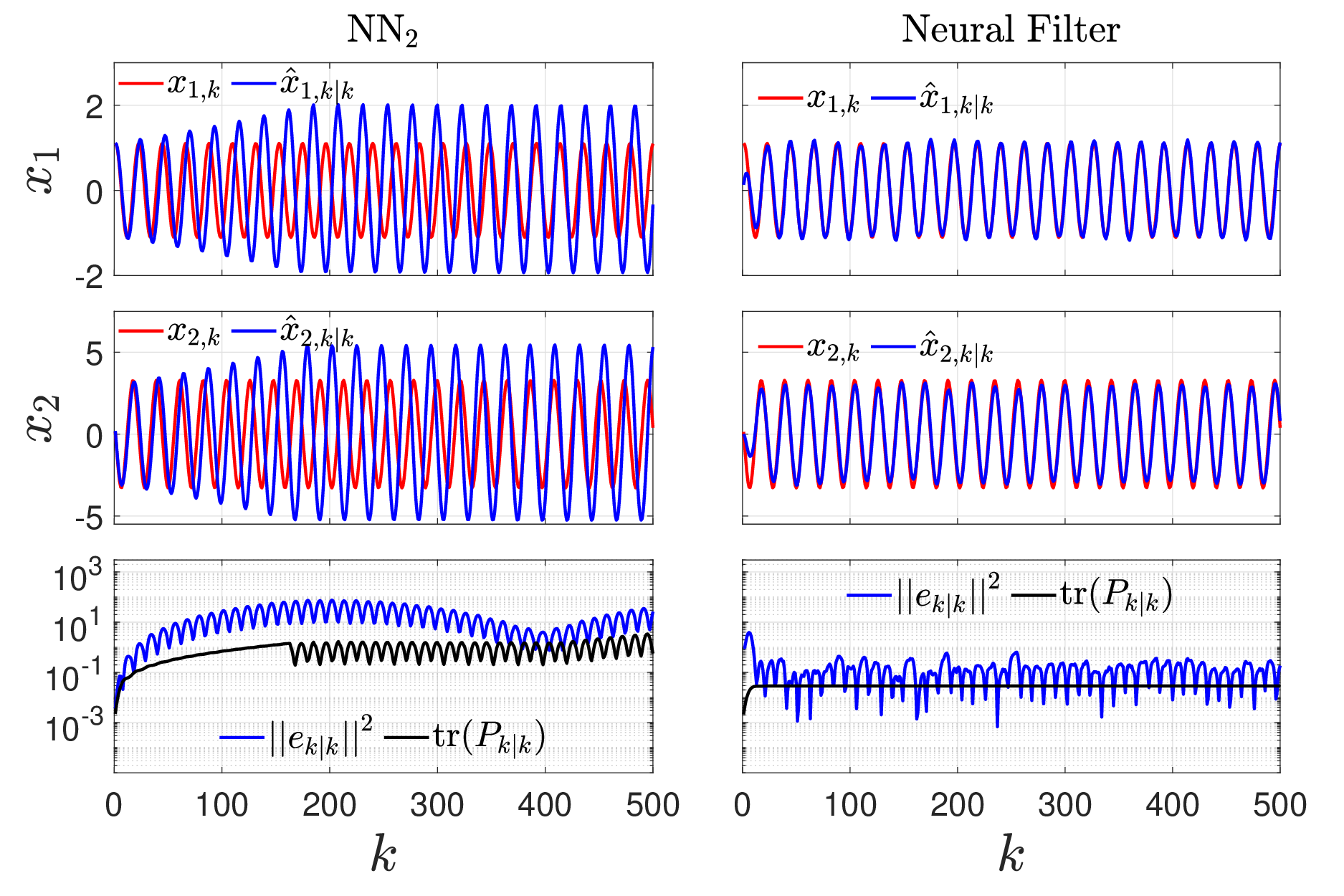}
    \caption{\textbf{Simple pendulum.} State predictions using the neural network and the neural filter using $\rm{NN}_2$. 
    }
    \label{fig:Pendulum_NN_NNEKF_fEKF_v2_bad}
\end{figure}

Next, the trained neural networks and the corresponding neural filters are used to predict the state of the pendulum system. 
In particular, we set 
$
    x(0)
        =
            \matl
                \frac{\pi}{3} &
                1
            \matr^\rmT. 
$
In both of the neural filter, we set $P(0) = 10^{-4} \times I_2,$ where $I_2$ is the 2 by 2 identity matrix and $\hat x_{0|0} = 0$.
Note that $\hat x_{0|0} = 0$ reflects the absence of knowledge of the state of the dynamic system in the neural filter. 
However, since the neural network models $NN_1$ and $NN_2$ have no external correction, we initialize them with the same initial conditions as the true system, that is, $\hat x_0 = 0$ in the neural network model.  
Figures \ref{fig:Pendulum_NN_NNEKF_fEKF_v2} and \ref{fig:Pendulum_NN_NNEKF_fEKF_v2_bad} show the predicted states using the neural network (subfigures in the left column) and the corresponding neural filter (subfigures in the right column) using $\rm{NN}_1$ and $\rm NN_2,$ respectively.

Note that, in both cases, the state predictions using the neural network degrade over time, as shown by the increasing state error norm $\| e_{k|k}\| $ and the trace of the corresponding state covariance $\tr P_{k|k}.$
On the other hand, the state predictions using the neural filter remain accurate over time, as shown by the bounded state error norm $\| e_{k|k}\| $ and the trace of the corresponding state covariance $\tr P_{k|k}.$
Furthermore, the $NN_1$ and $NN_2$ estimates diverge even with the correct initialization, whereas the neural filter estimates converge despite the complete lack of the system's initial state. 
%
%
\textit{Although $\rm{NN}_2$ is a much simpler model and shows noticeably poorer performance than $\rm{NN}_1$ in open-loop state estimation, the neural filter in both instances operates quite similarly.}





\subsection{Van Der Pol}
Consider the Van Der Pol oscillator
\begin{align}
    \label{eq:VDP_ODE} 
     \ddot q -\mu (1 - q^2) \dot q + q=
    0.
\end{align}
Defining
$
    x
        \isdef
            \matl
                q &
                \dot q
            \matr^\rmT, 
$
the Van der Pol oscillator \eqref{eq:VDP_ODE} can be written in the state-space form \eqref{eq:ss_form}, 
where 
\begin{align}
    f(x)
        \isdef
            \matl 
                x_2 \\
                \mu (1 - {x_1}^2)x_2 - x_1
            \matr.
\end{align}

The measurement model for this example is defined as follows
\begin{align}
    y_k = \begin{bmatrix}
       1  & 0
    \end{bmatrix}x_k + v_k, 
\end{align}
where $v_k \sim \SN(0, \sigma_v^2 I_2)$ represents zero-mean Gaussian noise. For this example, we specify $\sigma_v = 0.1$.

The training data consists of 30,000 samples of $x(0) \in \matl \SU[-5, 5] \\ \SU[-5, 5] \matr.$
Using MATLAB's \href{https://www.mathworks.com/help/matlab/ref/ode45.html}{ode45} routine, $x(0.1)$ is computed according to \eqref{eq:x_t_SP1}. In this work, we use 80 $\%$ of the data to train the model, and the remaining 20 $\%$ are used for validation during the training process.

\begin{figure}[h]
    \centering
    \includegraphics[width=\columnwidth]{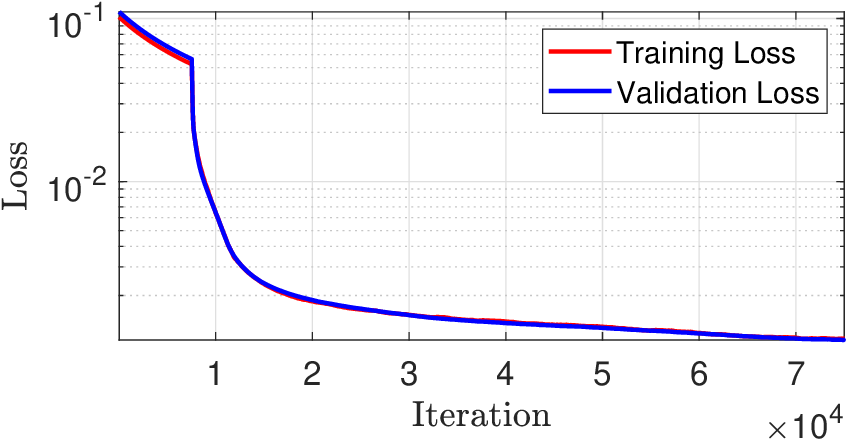}
    \caption{\textbf{Van der Pol oscillator.} Smoothed training and validation loss on logarithmic scale during the training process.}
    \label{fig:VDP_training_process}
\end{figure}

The neural network architecture used to approximate the Van der Pol oscillator is shown in Figure \ref{fig:VDP NN structure}.
In particular, the neural network consists of an input layer with a dimension of 2, two hidden layers containing 10 neurons each, and an output layer with a dimension of 2. 
Both hidden layers use the rectified linear unit (ReLU) activation function, while the output layer uses a linear activation function. 
The Adam optimizer is used for training the neural network.
Training is performed with a batch size of 32, and validation is conducted every 30 iterations.
Figure \ref{fig:VDP_training_process} shows the smoothed training and validation loss on logarithmic scale during the training process.

\begin{figure}[ht]
	\centering
        \begin{tikzpicture}[scale=1.2, transform shape]

\node[circle, draw, minimum size=0.3cm] (I-1) at (0,2) {};
\node[anchor=east] at (I-1.west) {$x_1(0)$};

\node[circle, draw, minimum size=0.3cm] (I-2) at (0,0) {};
\node[anchor=east] at (I-2.west) {$x_2(0)$};

\foreach \i in {1,...,10} {
    \node[circle, draw, minimum size=0.3cm] (H1-\i) at (1,3.75-\i*0.5) {};
}

\foreach \i in {1,...,10} {
    \node[circle, draw, minimum size=0.3cm] (H2-\i) at (3,3.75-\i*0.5) {};
}

\node[circle, draw, minimum size=0.3cm] (O-1) at (4,2) {};
\node[anchor=west] at (O-1.east) {$x_1(0.1)$};

\node[circle, draw, minimum size=0.3cm] (O-2) at (4,0) {};
\node[anchor=west] at (O-2.east) {$x_2(0.1)$};

\foreach \i in {1,2} {
    \foreach \j in {1,...,10} {
        \draw[->] (I-\i) -- (H1-\j);
    }
}

\foreach \i in {1,...,10} {
    \foreach \j in {1,...,10} {
        \draw[->] (H1-\i) -- (H2-\j);
    }
}

\foreach \i in {1,...,10} {
    \draw[->] (H2-\i) -- (O-1);
    \draw[->] (H2-\i) -- (O-2);
}

\end{tikzpicture}
 		
    \caption{Neural network architecture used to approximate the Van der Pol oscillator.}
    \label{fig:VDP NN structure}
\end{figure}

\begin{figure}[h]
    \centering
    \includegraphics[width=\columnwidth, trim={0.5cm 0 0.8cm 0},clip]{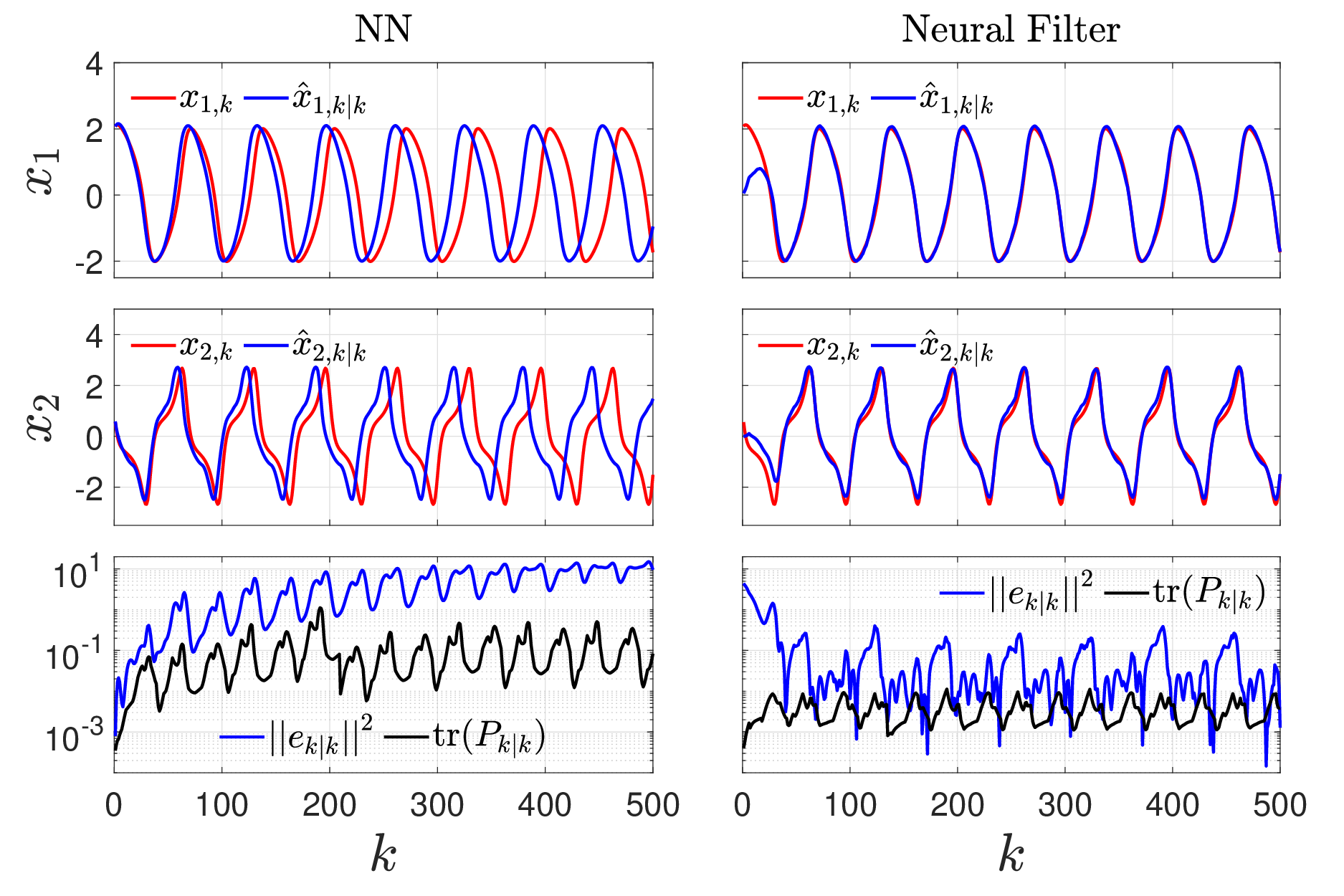}
    \caption{
    \textbf{Van der Pol oscillator.} State predictions using the neural network and the neural filter. }
    \label{fig:VDP_NN_NNEKF_fEKF_v2}
\end{figure}

Next, the trained neural network and the neural filter are used to predict the state of the Van der Pol oscillator. 
In particular, we set
$
    x(0)
        =
            \matl
                2 &
                1
            \matr^\rmT. 
$
In the neural filter, we set $P(0) = 10^{-4} \times I_2,$ where $I_2$ is the 2 by 2 identity matrix.
Figure \ref{fig:VDP_NN_NNEKF_fEKF_v2} shows the predicted states using the neural network (subfigures in the left column) and the neural filter (subfigures in the right column).
Note that the state predictions using the neural network degrade over time, as shown by the increasing state error norm $\| e_{k|k}\| $ and the trace of the corresponding state covariance $\tr P_{k|k}.$
On the other hand, the state predictions using the neural filter remain accurate over time, as shown by the bounded state error norm $\| e_{k|k}\| $ and the trace of the corresponding state covariance $\tr P_{k|k}.$



\subsection{Lorenz System}
Consider the Lorenz system
\begin{align}
    \label{eq: Lorenz_ODE}
     \dot w &= \sigma (y - w), \nn\\
     \dot y &= w(\rho - z) - y, \nn\\
     \dot z &= wy - \beta z.
\end{align}
Defining 
$
    x
        \isdef
            \matl
                w &
                y &
                z
            \matr^\rmT, 
$
the Lorenz system \eqref{eq: Lorenz_ODE} can be written in the state-space form \eqref{eq:ss_form}, where 
\begin{align}
    f(x)
        \isdef
            \matl 
                \sigma (x_2 - x_1) \\
                x_1(\rho - x_3) - x_2 \\
                x_1 x_2 - \beta x_3
            \matr.
\end{align}

The measurement model for this example is defined as follows
\begin{align}
    y_k = \begin{bmatrix}
       1  & 0 & 0
    \end{bmatrix}x_k + v_k, 
\end{align}
where $v_k \sim \SN(0, \sigma_v^2 I_3)$ represents zero-mean Gaussian noise. For this example, we specify $\sigma_v = 0.1$.

The training data consists of 100,000 samples of $x(0) \in \matl \SU[-15, 15] \\ \SU[-15, 15] \\ \SU[-15, 15] \matr.$
Using MATLAB's \href{https://www.mathworks.com/help/matlab/ref/ode45.html}{ode45} routine, $x(0.01)$ is computed according to \eqref{eq:x_t_SP1}. In this work, we use 80 $\%$ of the data to train the model, and the remaining 20 $\%$ are used for validation during the training process.

\begin{figure}[h]
    \centering
    \includegraphics[width=\columnwidth]{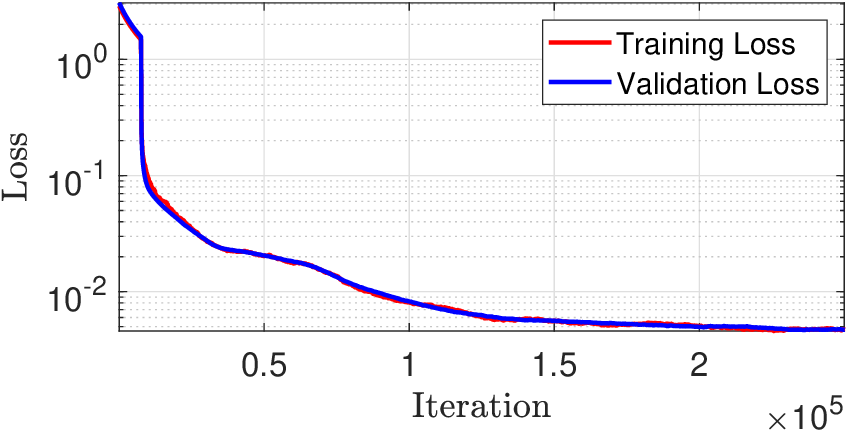}
    \caption{\textbf{Lorenz system.} Smoothed training and validation loss on logarithmic scale during the training process.}
    \label{fig:Lorenz_training_process}
\end{figure}

The neural network architecture used to approximate the Lorenz system is shown in Figure \ref{fig:Lorenz NN structure}.
In particular, the neural network consists of an input layer with a dimension of 3, three hidden layers containing 10 neurons each, and an output layer with a dimension of 3. 
All the hidden layers use the rectified linear unit (ReLU) activation function, while the output layer uses a linear activation function. 
The Adam optimizer is used for training the neural network.
Training is performed with a batch size of 32, and validation is conducted every 30 iterations.
Figure \ref{fig:Lorenz_training_process} shows the smoothed training and validation loss on logarithmic scale during the training process.

\begin{figure}[ht]
	\centering
        \begin{tikzpicture}[scale=1.2, transform shape]

\node[circle, draw, minimum size=0.3cm] (I-1) at (0,2) {};
\node[anchor=east] at (I-1.west) {$x_1(0)$};

\node[circle, draw, minimum size=0.3cm] (I-2) at (0,1) {};
\node[anchor=east] at (I-2.west) {$x_2(0)$};

\node[circle, draw, minimum size=0.3cm] (I-3) at (0,0) {};
\node[anchor=east] at (I-3.west) {$x_3(0)$};

\foreach \i in {1,...,10} {
    \node[circle, draw, minimum size=0.3cm] (H1-\i) at (1,3.75-\i*0.5) {};
}

\foreach \i in {1,...,10} {
    \node[circle, draw, minimum size=0.3cm] (H2-\i) at (2,3.75-\i*0.5) {};
}

\foreach \i in {1,...,10} {
    \node[circle, draw, minimum size=0.3cm] (H3-\i) at (3,3.75-\i*0.5) {};
}

\node[circle, draw, minimum size=0.3cm] (O-1) at (4,2) {};
\node[anchor=west] at (O-1.east) {$x_1(0.01)$};

\node[circle, draw, minimum size=0.3cm] (O-2) at (4,1) {};
\node[anchor=west] at (O-2.east) {$x_2(0.01)$};

\node[circle, draw, minimum size=0.3cm] (O-3) at (4,0) {};
\node[anchor=west] at (O-3.east) {$x_3(0.01)$};

\foreach \i in {1,2,3} {
    \foreach \j in {1,...,10} {
        \draw[->] (I-\i) -- (H1-\j);
    }
}

\foreach \i in {1,...,10} {
    \foreach \j in {1,...,10} {
        \draw[->] (H1-\i) -- (H2-\j);
    }
}

\foreach \i in {1,...,10} {
    \foreach \j in {1,...,10} {
        \draw[->] (H2-\i) -- (H3-\j);
    }
}

\foreach \i in {1,...,10} {
    \draw[->] (H3-\i) -- (O-1);
    \draw[->] (H3-\i) -- (O-2);
    \draw[->] (H3-\i) -- (O-3);
}

\end{tikzpicture}
 		
    \caption{Neural network architecture used to approximate the Lorenz system.}
    \label{fig:Lorenz NN structure}
\end{figure}

\begin{figure}[h]
    \centering
    \includegraphics[width=\columnwidth, trim={0.5cm 0 0.65cm 0},clip]{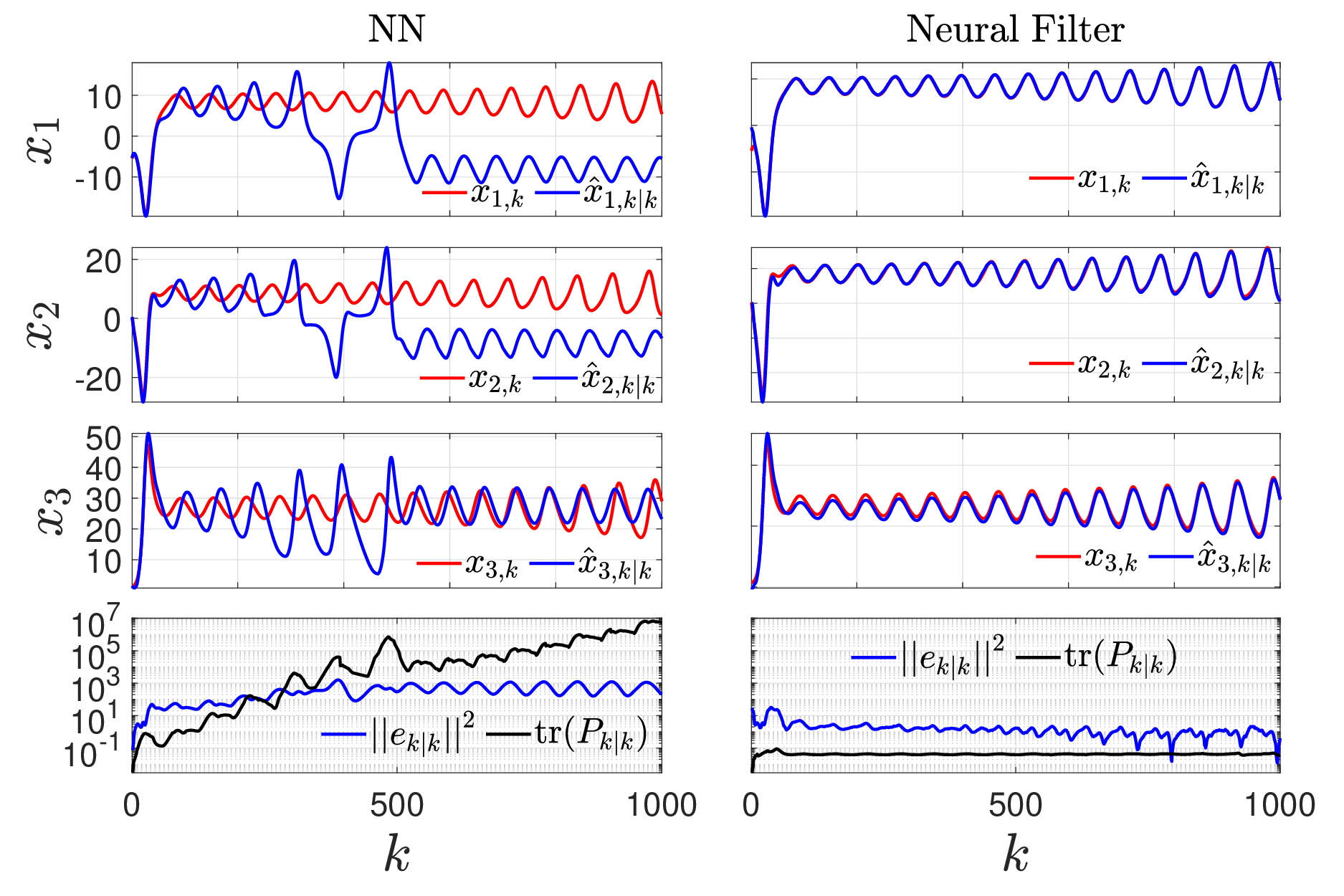}
    \caption{
    \textbf{Lorenz system.} State predictions using the neural network and the neural filter. }
    \label{fig:Lorenz_NN_NNEKF_fEKF_v2}
\end{figure}

Next, the trained neural network and the neural filter are used to predict the state of the Lorenz system. 
In particular, we set 
$
    x(0)
        =
            \matl
                -6.13 &
                1.78 &
                1.67
            \matr^\rmT. 
$
In the neural filter, we set $P(0) = 10^{-4} \times I_3,$ where $I_3$ is the 3 by 3 identity matrix.
Figure \ref{fig:Lorenz_NN_NNEKF_fEKF_v2} shows the predicted states using the neural network (subfigures in the left column) and the neural filter (subfigures in the right column).
Note that the state predictions using the neural network degrade over time, as shown by the increasing state error norm $\| e_{k|k}\| $ and the trace of the corresponding state covariance $\tr P_{k|k}.$
On the other hand, the state predictions using the neural filter remain accurate over time, as shown by the bounded state error norm $\| e_{k|k}\| $ and the trace of the corresponding state covariance $\tr P_{k|k}.$




\subsection{Double Inverted Pendulum System}

As shown in Figure \ref{Planardoublependulumfigure}, a planar double simple pendulum consists of particles $y_1$ and $y_2$ with masses $m_1$ and $m_2,$ respectively.
The particle $y_1$ is connected to a frictionless pin joint at the point $w$ in the ceiling by means of the massless link $\SL_1$ with length $\ell_1,$
and the particle $y_2$ is connected by a frictionless pin joint at $y_1$ to the first link by means of the massless link $\SL_2$ with length $\ell_2.$
The external torque $\tau_{\rm ext}$ is applied to link $\SL_1.$
%

\begin{figure}[H]
	\centering
 		\begin{tikzpicture}[auto, node distance=2cm,>=latex']
            \tikzmath{
            \x0=0; \y0=0;
            \x1=3*cos(-120);
            \y1=3*sin(-120);
            \x2=3*cos(-120)+3*cos(-140); 
            \y2=3*sin(-120)+3*sin(-140);
            }
            
            \node (link1) at (\x1+.5,\y1+1.5) {$\SL_1$}; 
            \node (link2) at (\x2+.75,\y2+1) {$\SL_2$}; 
            
            \draw [dashed, thick] (\x0, \y0) -- +(0, -2);
            \draw [dashed, thick] (\x1, \y1) -- +(0, -2);
            \draw [thick] (\x0, \y0) -- +(2, 0)-- +(-2, 0);
            
            \draw [line width = 1] (\x0, \y0) 
                            -- (\x1, \y1)
                            -- (\x2, \y2);
            

            \draw [thick, fill=black] (\x0,\y0) 
                    node[xshift=0, yshift=10] {$w$}
                    circle [radius=.05];
		    \draw [thick, fill=black] (\x1,\y1)
                    node[xshift=10, yshift=0] {$y_1$}
                    circle [radius=.1];
		    \draw [thick, fill=black] (\x2,\y2) 
                    node[xshift=10, yshift=0] {$y_2$}
                    circle [radius=.1];
                            
            \draw [thick,->] (0,-1) 
                    node[xshift=-10, yshift=-7] {$\theta_1$} 
                    arc (-90:-120:1);
            \draw [thick,->] ({3*cos(-120)},{3*sin(-120)-1}) 
                    node[xshift=-15, yshift=-7] {$\theta_2$} 
                    arc (-90:-140:1);
            

            \draw [ultra thick,->] (0,-3) -- +(0,-1)     
                    node[xshift=10,yshift=10] {$\vect g$};
            
		\end{tikzpicture}
        \caption{\small{
        Planar double simple pendulum.  
        }}
        \label{Planardoublependulumfigure}
\end{figure}

The equations of motion of the double pendulum are
\begin{align}
            (m_2+ m_1) \ell_1^2 \ddot \theta_1 
            &+
            m_2 \ell_1 \ell_2 (\cos \phi)\ddot\theta_2             
                =            
                     m_2 \ell_1 \ell_2 (\sin \phi)\dot\theta_2^2 
                    \nn \\ &\quad 
                    - (m_1  + m_2)g \ell_1 \sin \theta_1
                    +\tau_{\rm ext},
            \\
        \ell_1 (\cos \phi)\ddot\theta_1 
        &+
        \ell_2 \ddot\theta_2 
        =             
            - \ell_1 (\sin \phi)\dot \theta_1^2 
            -g \sin \theta_2 ,
\end{align}
where $\phi \isdef \theta_2 - \theta_1.$

The equations of motion can be written compactly as
\begin{align}
    \SM(\theta) \ddot \theta + \SD(\theta, \dot \theta)
        = T, 
    \label{eq:NLS}
\end{align}
where $\theta \isdef \matl \theta_1 \\ \theta_2 \matr$ and 
\begin{align}
    \SM(\theta)
        &\isdef
            \matl 
                (m_2+ m_1) \ell_1^2 & m_2 \ell_1 \ell_2 (\cos \phi) \\
                \ell_1 (\cos \phi) & \ell_2 
            \matr, 
    \\
    \SD(\theta, \dot \theta)
        &\isdef
            \matl 
                -m_2 \ell_1 \ell_2 (\sin \phi)\dot\theta_2^2 
                + (m_1  + m_2)g \ell_1 \sin \theta_1 \\
                \ell_1 (\sin \phi)\dot \theta_1^2 
                +g \sin \theta_2 
            \matr, 
    \\ 
    T 
        &\isdef
            \matl 
                \tau_{\rm ext} \\
                0
            \matr.
\end{align}

The measurement model for this example is defined as follows
\begin{align}
    y_k = \begin{bmatrix}
       1  & 0 & 0 & 0\\
       0  & 0 & 1 & 0
    \end{bmatrix}x_k + v_k, 
\end{align}
where $v_k \sim \SN(0, \sigma_v^2 I_4)$ represents zero-mean Gaussian noise. For this example, we specify $\sigma_v = 0.01$.

The training data consists of 200,000 samples of $x(0) \in \matl \SU[-\pi/2, \pi/2] \\ \SU[-0.5, 0.5] \\ \SU[-\pi/2, \pi/2] \\ \SU[-0.5, 0.5]  \matr.$
Using MATLAB's \href{https://www.mathworks.com/help/matlab/ref/ode45.html}{ode45} routine, $x(0.01)$ is computed according to \eqref{eq:x_t_SP1} assuming $ \tau_{\rm ext} = 0$. In this work, we use 80 $\%$ of the data to train the model, and the remaining 20 $\%$ are used for validation during the training process.

\begin{figure}[h]
    \centering
    \includegraphics[width=\columnwidth]{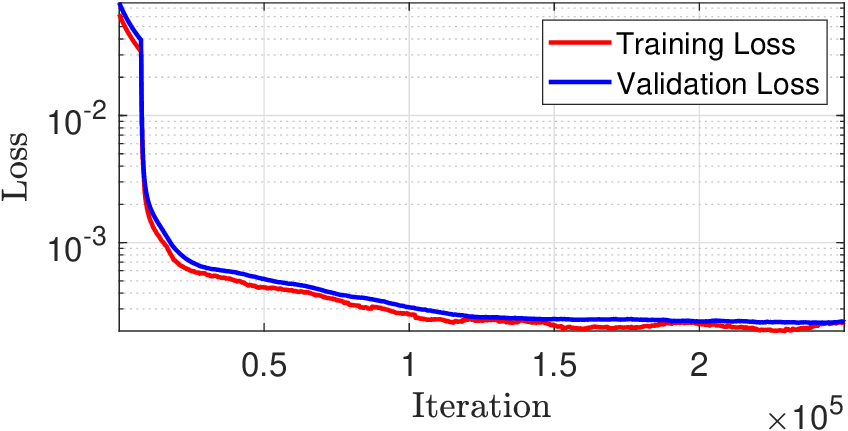}
    \caption{\textbf{Double pendulum.} Smoothed training and validation loss on logarithmic scale during the training process.}
    \label{fig:Double_Pendulum_training_process}
\end{figure}

The neural network architecture used to approximate the double pendulum system is shown in Figure \ref{fig:double pendulum structure}.
In particular,  the neural network consists of an input layer with a dimension of 4, four hidden layers containing 10 neurons each, and an output layer with a dimension of 4.
All the hidden layers use the rectified linear unit (ReLU) activation function, while the output layer uses a linear activation function. 
The Adam optimizer is used for training the neural network.
Training is performed with a batch size of 32, and validation is conducted every 30 iterations.
Figure \ref{fig:Double_Pendulum_training_process} shows the smoothed training and validation loss on a logarithmic scale during the training process.

\begin{figure}[ht]
	\centering
        \begin{tikzpicture}[scale=1.2, transform shape]

\node[circle, draw, minimum size=0.3cm] (I-1) at (0,3) {};
\node[anchor=east] at (I-1.west) {$x_1(0)$};

\node[circle, draw, minimum size=0.3cm] (I-2) at (0,2) {};
\node[anchor=east] at (I-2.west) {$x_2(0)$};

\node[circle, draw, minimum size=0.3cm] (I-3) at (0,1) {};
\node[anchor=east] at (I-3.west) {$x_3(0)$};

\node[circle, draw, minimum size=0.3cm] (I-4) at (0,0) {};
\node[anchor=east] at (I-4.west) {$x_4(0)$};

\foreach \i in {1,...,10} {
    \node[circle, draw, minimum size=0.3cm] (H1-\i) at (0.6,4.25-\i*0.5) {};
}

\foreach \i in {1,...,10} {
    \node[circle, draw, minimum size=0.3cm] (H2-\i) at (1.6,4.25-\i*0.5) {};
}

\foreach \i in {1,...,10} {
    \node[circle, draw, minimum size=0.3cm] (H3-\i) at (2.6,4.25-\i*0.5) {};
}

\foreach \i in {1,...,10} {
    \node[circle, draw, minimum size=0.3cm] (H4-\i) at (3.6,4.25-\i*0.5) {};
}

\node[circle, draw, minimum size=0.3cm] (O-1) at (4.2,3) {};
\node[anchor=west] at (O-1.east) {$x_1(0.01)$};

\node[circle, draw, minimum size=0.3cm] (O-2) at (4.2,2) {};
\node[anchor=west] at (O-2.east) {$x_2(0.01)$};

\node[circle, draw, minimum size=0.3cm] (O-3) at (4.2,1) {};
\node[anchor=west] at (O-3.east) {$x_3(0.01)$};

\node[circle, draw, minimum size=0.3cm] (O-4) at (4.2,0) {};
\node[anchor=west] at (O-4.east) {$x_4(0.01)$};

\foreach \i in {1,2,3,4} {
    \foreach \j in {1,...,10} {
        \draw[->] (I-\i) -- (H1-\j);
    }
}

\foreach \i in {1,...,10} {
    \foreach \j in {1,...,10} {
        \draw[->] (H1-\i) -- (H2-\j);
    }
}

\foreach \i in {1,...,10} {
    \foreach \j in {1,...,10} {
        \draw[->] (H2-\i) -- (H3-\j);
    }
}

\foreach \i in {1,...,10} {
    \foreach \j in {1,...,10} {
        \draw[->] (H3-\i) -- (H4-\j);
    }
}

\foreach \i in {1,...,10} {
    \draw[->] (H4-\i) -- (O-1);
    \draw[->] (H4-\i) -- (O-2);
    \draw[->] (H4-\i) -- (O-3);
    \draw[->] (H4-\i) -- (O-4);
}

\end{tikzpicture}
 		
    \caption{Neural network architecture used to approximate the double pendulum system.}
    \label{fig:double pendulum structure}
\end{figure}

\begin{figure}[h]
    \centering
    \includegraphics[width=\columnwidth, trim={0.5cm 0 0.65cm 0},clip]{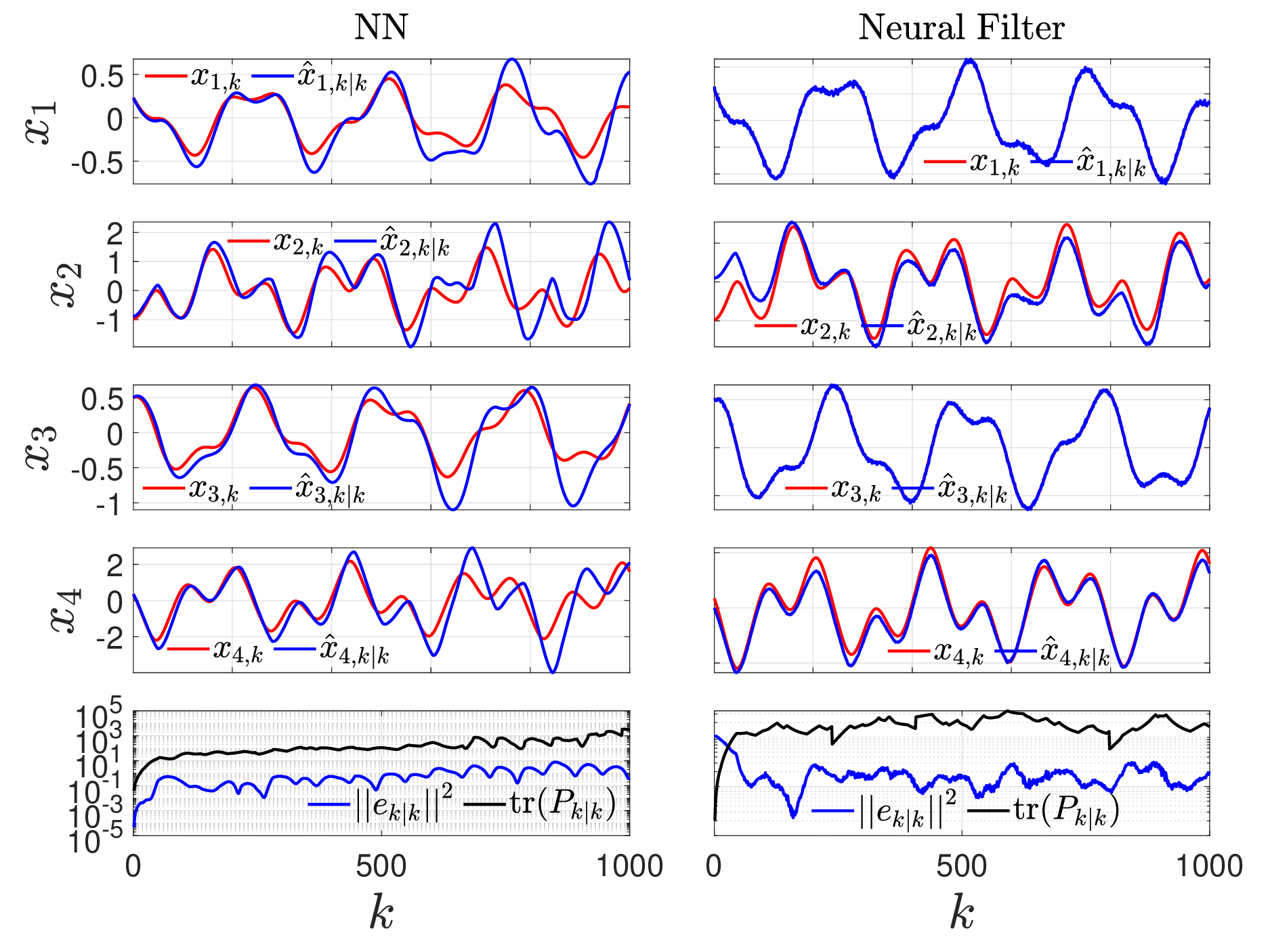}
    \caption{
    \textbf{Double pendulum.} State predictions using the neural network and the neural filter.}
    \label{fig:Double_Pendulum_NN_NNEKF_fEKF_v2}
\end{figure}

Next, the trained neural network and the neural filter are used to predict the state of the double pendulum system. 
In particular, we set 
$
    x(0)
        =
            \matl
                -0.235 &
                0.267 &
                -0.435 &
                -0.301
            \matr^\rmT. 
$
In the neural filter, we set $P(0) = 10^{-4} \times I_4,$ where $I_4$ is the 4 by 4 identity matrix.
Figure \ref{fig:Double_Pendulum_NN_NNEKF_fEKF_v2} shows the predicted states using the neural network (subfigures in the left column) and the neural filter (subfigures in the right column).
Note that the state predictions using the neural network degrade over time, as shown by the increasing state error norm $\| e_{k|k}\| $ and the trace of the corresponding state covariance $\tr P_{k|k}.$
On the other hand, the state predictions using the neural filter remain accurate over time, as shown by the bounded state error norm $\| e_{k|k}\| $ and the trace of the corresponding state covariance $\tr P_{k|k}.$

\section{Conclusions}
\label{sec:Conclusions}
This paper introduced a novel neural filter to improve the accuracy of state predictions using neural network-based models of dynamical systems. 
Motivated by the extended Kalman filter, the neural filter is constructed using the neural network-based approximation of the dynamical system and augmenting the state prediction by a correction step.
The performance of the neural filter is investigated using four nonlinear dynamical systems.
In particular, a simple pendulum, a Van der Pol oscillator, Lorenz system, and a double pendulum system are considered to demonstrate the application of the neural filter. 
In each case, the states predicted by the neural filter remain close to the true states, whereas the state estimates provided by the neural network model alone diverge. 
%
Moreover, unlike the neural network model, which requires precise initialization, the neural filter state estimates latch onto true states even with zero initialization. 
%
%
Furthermore, the covariance of the state estimate given by the neural filter remains bounded, whereas the covariance of the state estimate given by the neural network model diverges.

\section{Appendix}
\subsection{Jacobian of the Neural Network}
\label{appdx:jacobian_NN}

Following the notation presented in \cite{rozario2022tutorial}, the output of the neural network is given by
\begin{align}
    y &= NN(x, \Theta_1, \ldots, \Theta_n,\Theta_{n+1})
        \nn \\
        = 
            &\Theta_{n+1}^\rmT N_n ( N_{n-1} (\ldots ( N_1(x,\Theta_1)),\Theta_{n-1} ),\Theta_n ),
\end{align}
where $N_1, N_2, \ldots, N_n$ are the $n$ neural layers of the neural network and $\{ \Theta_i \}_{i=1}^n$ are the neural gains in each layer. 

To compute the Jacobian of the neural network function $NN$ with respect to the input $x,$ the output of a neural network function is written using a recursive formula as shown below. 
By denoting the input and the output of the $i$th neural layer by $x_i$ and $x_{i+1},$ it follows that, for $i \in \{1, 2, \ldots, n\},$ 
\begin{align}
    x_{i+1} = N_i(x_i, \Theta_i),
\end{align}
where $\Theta_i \in \BBR^{l_{\theta_i} \times \ell_i}.$
Note that $x_1 \isdef x.$
The output of the network is finally given by
\begin{align}
    y = S(x_{n+1}) =  \Theta_{n+1}^\rmT x_{n+1},
\end{align}
where $\Theta_{n+1} \in \BBR^{l_{x_{n+1}} \times l_y}.$
Using the chain rule, it follows that 
\begin{align}
    \frac{\partial NN}{\partial x} 
        &=
            \frac{\partial y}{\partial x} 
        =
            \frac{\partial y}{\partial x_{i+1}}
            \frac{\partial x_{i+1}}{\partial x_{i}}
            \frac{\partial x_{i}}{\partial x_{i-1}}
            \ldots 
            \frac{\partial x_{2}}{\partial x_1}
        \nn \\
        &=
            \Theta_{n+1}^\rmT
            \frac{\partial N_i}{\partial x_{i}}
            \ldots 
            \frac{\partial N_{1}}{\partial x_1}.
\end{align}
More details about computing the gradients of arbitrarily connected neural networks can be found in \cite{wilamowski2008computing}.

\printbibliography

\end{document}